\documentclass[a4paper]{article}

\usepackage{INTERSPEECH2022}
\usepackage{xcolor}
\usepackage{multirow}
\usepackage{pifont}
\usepackage{amsmath}
\usepackage[inline]{enumitem}
\usepackage{hyperref}
\usepackage{cleveref}[2012/02/15]
\usepackage{cite}

\crefformat{footnote}{#2\footnotemark[#1]#3}

\renewcommand*{\thefootnote}{\fnsymbol{footnote}}

\title{Enhanced Direct Speech-to-Speech Translation Using Self-supervised Pre-training and Data Augmentation}
\name{Sravya Popuri$^{\star}$, Peng-Jen Chen$^{\star}$, Changhan Wang, Juan Pino, \\ Yossi Adi, Jiatao Gu, Wei-Ning Hsu$^{\dagger}$, Ann Lee$^{\dagger}$}

\address{
  Meta AI}
\email{\{spopuri,pipibjc\}@fb.com}

\begin{document}

\maketitle

\renewcommand{\thefootnote}{$^{\star}$}
\footnotetext[1]{Equal contribution. $^{\dagger}$ Equal supervision.}
\renewcommand*{\thefootnote}{\arabic{footnote}}
\setcounter{footnote}{0}

\begin{abstract}
Direct speech-to-speech translation (S2ST) models suffer from data scarcity issues as there exists little parallel S2ST data, compared to the amount of data available for conventional cascaded systems that consist of automatic speech recognition (ASR), machine translation (MT), and text-to-speech (TTS) synthesis.
In this work, we explore self-supervised pre-training with unlabeled speech data and data augmentation to tackle this issue.
We take advantage of a recently proposed speech-to-unit translation (S2UT) framework that encodes target speech into discrete representations, and transfer pre-training and efficient partial finetuning techniques that work well for speech-to-text translation (S2T) to the S2UT domain by studying both speech encoder and discrete unit decoder pre-training. Our experiments on Spanish-English translation show that self-supervised pre-training consistently improves model performance compared with multitask learning with an average 6.6-12.1 BLEU gain, and it can be further combined with data augmentation techniques that apply MT to create weakly supervised training data.\footnote{Audio samples are available at: \url{https://facebookresearch.github.io/speech\_translation/enhanced\_direct\_s2st\_units/index.html}. Code and pre-trained models are available at: \url{https://github.com/pytorch/fairseq/blob/main/examples/speech\_to\_speech/docs/enhanced\_direct\_s2st\_discrete\_units.md}}

\end{abstract}

\section{Introduction}
Direct speech-to-speech translation (S2ST) aims at translating speech from one language into speech in another language without relying on text generation as an intermediate step~\cite{jia2019direct,tjandra2019speech,zhang2020uwspeech,lee2021direct,jia2021translatotron,lee2021textless}.
Compared to conventional cascaded approaches~\cite{lavie1997janus,nakamura2006atr}, which take advantage of automatic speech recognition (ASR), machine translation (MT) or end-to-end speech-to-text translation (S2T) followed by text-to-speech synthesis (TTS), direct S2ST has the advantage of faster inference~\cite{lee2021direct} and can support translation between languages without text writing systems~\cite{tjandra2019speech,zhang2020uwspeech,lee2021direct,lee2021textless}.

Most recently,~\cite{lee2021direct} proposes to apply a self-supervised speech encoder pre-trained on unlabeled speech to convert target speech into discrete units~\cite{lakhotia2021generative} and build a speech-to-unit translation (S2UT) model for direct S2ST. Self-supervised discrete targets can disentangle linguistic content from speaker identity and prosodic information in speech~\cite{polyak2021speech}. Moreover, they enable opportunities for applying techniques from speech-to-text model training, such as ASR and S2T, to direct S2ST. 

As we move along the spectrum from multi-stage to direct approaches, the amount of parallel data available becomes much more scarce.
Pre-training, including initializing encoder or decoder trained from ASR or MT tasks, or from self-supervised pre-training with unlabeled data~\cite{berard2018end,bahar2019comparative,stoian2020analyzing,li2021multilingual}, multitask learning~\cite{weiss2017sequence,berard2018end} and data augmentation~\cite{jia2019leveraging,stoian2020analyzing,pino2020self,wang2021large} have been extensively studied for tackling the data scarcity issue for S2T. For direct S2ST,~\cite{jia2019direct,lee2021direct} show that multitask learning is crucial for model convergence, and~\cite{jia2019direct,kano2021transformer} focus on incorporating pre-trained modules from ASR, S2T, MT or TTS tasks.

In this work, we take advantage of the S2UT framework proposed in~\cite{lee2021direct} and show that large-scale self-supervised pre-training with monolingual speech and text data and data augmentation techniques that benefit S2T model training can also be applied on S2UT model training.
For pre-training, we transfer the technique from~\cite{li2021multilingual}, which performs efficient finetuning with a wav2vec 2.0 speech encoder~\cite{baevski2020wav2vec} and an mBART text decoder~\cite{liu2020multilingual}, to S2UT with a wav2vec 2.0 speech encoder and an mBART decoder trained with discrete units extracted from unlabeled speech data.
For data augmentation, we utilize ASR, MT and TTS models to create weakly supervised data~\cite{jia2019leveraging}.

The contributions of this work are as follows: we empirically demonstrate that self-supervised encoder and decoder pre-training with unlabeled speech and partial finetuning improve S2ST training under various setups, including training with synthetic single-speaker or real multi-speaker target speech, and low-resource setup (30-hr). We further improve the model via data augmentation with weakly supervised data and conduct experiments with the combination of multiple datasets to achieve strong performance over ASR+MT+TTS baseline systems.

\section{Related Work}
Self-supervised speech encoder pre-training has led to huge performance improvement on a wide range of applications~\cite{yang2021superb} such as ASR~\cite{baevski2020wav2vec,hsu2021hubert}, S2T~\cite{wang2021large}, speaker identification~\cite{chen2021wavlm}, etc. SpeechT5~\cite{ao2021speecht5} proposes end-to-end pre-training, where the model learns to reconstruct the log Mel-filterbank of the masked regions of the input speech, to improve performance of speech-to-speech tasks such as voice conversion and speech enhancement, while it was not evaluated in the context of direct S2ST.

The improved quality of speech units discretized based on self-supervised speech representations also allows researchers to apply natural language processing (NLP) techniques on speech, such as spoken generative language modeling~\cite{lakhotia2021generative,kharitonov2021text}, and emotion conversion cast as a unit-to-unit translation task~\cite{kreuk2021textless}.~\cite{kreuk2021textless} further applies text-based autoencoder denoising on units for model pre-training. In this work, we model the target speech as discrete units, extend the monolingual unit pre-training in~\cite{kreuk2021textless} to a multilingual setup, and perform speech encoder and discrete unit decoder pre-training separately.

Data augmentation is a common method for increasing the size of training data. Synthetic speech data from TTS~\cite{jia2019leveraging,tjandra2018machine}, self-training~\cite{wang2021large,kahn2020self} or back-translation~\cite{hayashi2018back} from unlabeled data have been shown to be useful to ASR or S2T.
In this work, we apply MT and TTS to prepare weakly supervised S2ST data from speech in the source language.

\vspace{-2mm}
\section{System}

\begin{figure}[t!]
  \centering
  \includegraphics[width=0.95\linewidth]{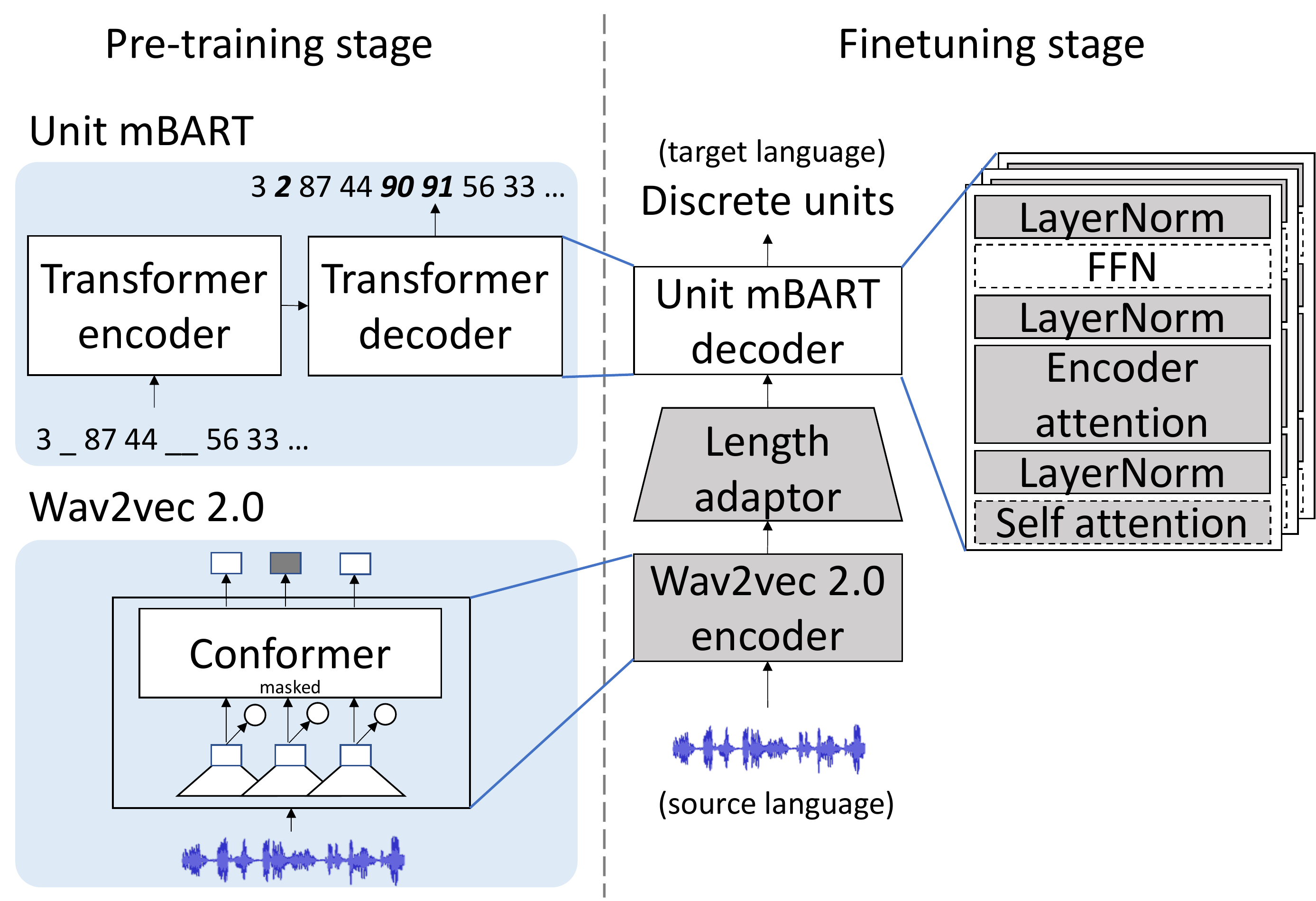}
  \caption{Flowchart for the speech encoder and decoder pre-training and finetuning process. The blocks in shade illustrate the modules that are finetuned in the \textbf{LNA-D} finetuning strategy (Sec.~\ref{sec:finetune}), which we find is the most effective. }
  \label{fig:flowchart}
  \vspace{-0.5cm}
\end{figure}

\subsection{Speech-to-unit translation (S2UT) model}

We follow~\cite{lee2021direct} and encode the target speech as discrete units with a HuBERT model trained on unlabeled speech followed by a k-means model learned from its hidden representations~\cite{hsu2021hubert}. Waveform input is encoded into a sequence of k-means cluster indices at every 20-ms frame, and we remove consecutive duplicate units to create a \textit{reduced} unit sequence representing the target speech. In the end, the direct S2ST system consists of a sequence-to-sequence S2UT model with a speech encoder and a unit decoder, followed by a unit HiFi-GAN vocoder~\cite{polyak2021speech} trained separately for unit-to-waveform conversion. In this work, we explore both encoder and decoder pre-training (Fig.~\ref{fig:flowchart}).

\vspace{-2mm}
\subsection{Model pre-training} 

\subsubsection{Encoder pre-training: wav2vec 2.0}
Wav2vec 2.0~\cite{baevski2020wav2vec} is a self-supervised framework to learn speech representations from unlabeled audio data. It uses a multi-layer convolution neural network to encode the audio followed by a Transformer-based~\cite{vaswani2017attention} context encoder to build the contextualized representations. The model is trained via contrastive loss with masked spans on the input to the context encoder. In this work, we use Conformer~\cite{gulati2020conformer} instead of the Transformer~\cite{vaswani2017attention} for better model performance (Sec.~\ref{sec:model_variation}).

\vspace{-0.25cm}
\subsubsection{Decoder pre-training: unit mBART}
mBART~\cite{liu2020multilingual} was originally proposed for denoising autoencoder over text.
During training, the sequence-to-sequence model predicts the original text $x$ given its noisy version, $g(x)$, created by randomly masking spans of $x$.
The starting position of each span is uniformly sampled from all positions, and the span lengths are sampled from a Poisson distribution ($\lambda$). The masking process is repeated until the total accumulated span lengths take up $p\%$ of the input. The model is trained on text data from multiple languages with language tags.
In our case, we treat the \textit{reduced} discrete units extracted from unlabeled speech data as text and apply mBART training with a Transformer-based encoder-decoder architecture~\cite{vaswani2017attention}.

\subsection{Model finetuning}
\label{sec:finetune}
We combine the wav2vec 2.0 encoder and the unit mBART decoder and study the finetuning strategies in~\cite{li2021multilingual}.
A randomly initialized adaptor layer consisting of a single 1-D convolutional layer with stride 2 is added between the pre-trained modules to increase the model's capacity to alleviate the mismatch between the learned representations, as well as length difference between the source audio and the \textit{reduced} target units. 

We examine both full and partial finetuning of the model. For the latter, we focus on the \textbf{L}ayer\textbf{N}orm and \textbf{A}ttention modules (dubbed as ``LNA'') proposed in~\cite{li2021multilingual}. The hypothesis is that LayerNorm parameters reflect the statistics of the pre-training data, and the encoder attention in the unit mBART decoder is optimized for unit sequence input. The adaptor layer is fully finetuned. We explore four finetuning strategies in total:
\begin{enumerate*}[leftmargin=*]
    \item \textbf{LNA-E}: The LayerNorm and self attention parameters in the encoder and all the parameters in the decoder are finetuned.
    \item \textbf{LNA-D}: The whole encoder and the LayerNorm and both encoder and self attention in the decoder are finetuned. We optionally freeze the encoder for the first $k$ updates.
    \item \textbf{LNA-E,D}: Only LNA parameters are finetuned both on the encoder and the decoder side. 
    \item \textbf{Full}: We finetune the whole model end-to-end with an option of freezing the encoder for the first $k$ updates.
\end{enumerate*}

\vspace{-2mm}
\subsection{Data augmentation}
We take advantage of speech from ASR data in the source language to increase the size of the parallel S2ST training data.
We use a Transformer MT model~\cite{vaswani2017attention} to translate the text transcription in the source language to text in the target language. To convert text in the target language into target speech, we apply a text-to-unit (T2U) model~\cite{lee2021textless}\footnote{\label{fn:textless}\url{https://github.com/pytorch/fairseq/blob/main/examples/speech_to_speech/docs/textless_s2st_real_data.md}}, which is a Transformer-based sequence-to-sequence model trained on text and the corresponding discrete unit sequence extracted from the paired audio. The T2U model is a way to bypass the TTS generation and HuBERT unit extraction pipeline for efficient generation of large-scale weakly supervised data. 
We choose to distill knowledge from MT models instead of pursuing self-training, since a three-stage cascaded system (ASR+MT+TTS) can take advantage of a large amount of data from each component during training and still outperforms existing direct S2ST systems~\cite{lee2021direct}.

\vspace{-2mm}
\section {Experiments}
We conduct experiments on Spanish-English (Es-En) and English-Spanish (En-Es) translation with \textsc{fairseq}~\cite{ott2019fairseq,wang2020fairseqs2t}.

\begin{table}[t!]
\caption{Statistics of datasets (train/dev/test splits). TTS is applied on the target text of S2T for creating synthetic S2ST data.
\vspace{-0.3cm}
}
\label{tab:dataset_stats_v2}
\centering
\resizebox{1.08\linewidth}{!}{
\begin{tabular}{l|ccc}
\hline
 & \# samples & source (hrs) & target (hrs) \\
\hline
\multicolumn{4}{l}{\textbf{S2T}, Es-En} \\
\hline
CoVoST-2~\cite{wang2020covost} & 78.9k / 13.3k / 13.2k & 112 / 22.0 / 22.7 & 81.0 / 14.4 / - \\
Europarl-ST~\cite{iranzo2020europarl} & 7.4k / 1.9k / 1.8k & 20.6 / 5.4 / 5.1 &  21.8 / 5.6 / - \\
mTEDx~\cite{salesky2021mtedx} & 35.6k / 888 / 989 & 63.4 / 1.5 / 1.7 & 58.6 / 1.4 / - \\
\hline
\multicolumn{4}{l}{\textbf{S2T}, En-Es} \\
\hline
Europarl-ST~\cite{iranzo2020europarl} & 31.6k / 1.3k / 1.3k & 75.6 / 3.0 / 2.9  &  76.5 / 3.0 / -\\
MuST-C~\cite{cattoni2021must} & 260k / 1.3k / 2.4k & 495 / 2.5 / 4.1 &  481 / 2.5 / - \\
\hline
\hline
\multicolumn{4}{l}{\textbf{S2ST}, Es-En} \\
\hline
VoxPopuli~\cite{wang2021voxpopuli,lee2021textless} & 159k / - / - & 532.1 / - / - & 513.1 / - / - \\
\hline
\multicolumn{4}{l}{\textbf{S2ST}, En-Es} \\
\hline
VoxPopuli~\cite{wang2021voxpopuli,lee2021textless} & 126k / - / - & 414.7 / - / - & 424.1 / - / - \\
\hline
\hline
\multicolumn{4}{l}{\textbf{ASR}, Es} \\
\hline
MLS~\cite{pratap2020mls} & 220k / 2.4k / 2.4k & 918 / 10.0 / 10.0 & - \\
CommonVoice 7.0~\cite{ardila2019common} & 196k / 15.3k / 15.3k & 290 / 25.7 / 26.2 & - \\
\hline
\multicolumn{4}{l}{\textbf{ASR}, En} \\
\hline
Librispeech~\cite{panayotov2015librispeech} & 282k / 5.6k / 5.5k & 960 / 10.5 / 10.7 & - \\
TED-LIUM3~\cite{hernandez2018ted} & 268k / 507 / 1.2k & 452 / 1.6 / 2.6 & - \\
\hline
\hline
\multicolumn{4}{l}{\textbf{Unlabeled Speech}, Es} \\
\hline
VoxPopuli~\cite{wang2021voxpopuli} & 2.0M & 16k & - \\
\hline
\multicolumn{4}{l}{\textbf{Unlabeled Speech}, En} \\
\hline
VoxPopuli~\cite{wang2021voxpopuli} & 1.8M & 14k & - \\
Librilight~\cite{kahn2020libri} & 18.6M & 60k & -\\
\hline
\hline
\multicolumn{4}{l}{\textbf{Parallel Text}, Es-En \& En-Es} \\
\hline
OpenSubtitle2018~\cite{lison-etal-2018-opensubtitles2018} & 64.7M & -  & - \\
UNCorpus~\cite{ziemski2016united} & 25M & - & -\\
 EUBookshop v2~\cite{skadins-etal-2014-billions} & 5.3M & - & -  \\
Europarl v10~\cite{koehn2005europarl} & 1.9M & - & - \\
Wikipedia v1.0~\cite{wolk2014building} & 1.8M & - & - \\
TED2020 v1~\cite{reimers-2020-multilingual-sentence-bert} & 0.4M & - & -  \\
\end{tabular}}
\vspace{-0.6cm}
\end{table}

\vspace{-2mm}
\subsection {Data}
Table~\ref{tab:dataset_stats_v2} summarizes the statistics of all the datasets used in the experiments.
We experiment with two types of parallel \textbf{S2ST data}. First, we follow the convention of applying single-speaker TTS\footnote{\label{fn:tts} En: \url{https://huggingface.co/facebook/tts\_transformer-en-ljspeech}, Es: \url{https://huggingface.co/facebook/tts\_transformer-es-css10}} on the target text of \textbf{S2T data}~\cite{jia2019direct,jia2021translatotron,lee2021direct,zhang2020uwspeech} (dubbed as ``\textit{S2ST-syn}''). We combine S2T datasets from multiple domains to improve the robustness of model training~\cite{likhomanenko2020rethinking,chan2021speechstew}, resulting in 196-hr training data for Es-En and 571-hr for En-Es. We also combine the dev sets from all domains for tracking the training process and checkpoint selection, and conduct evaluation on all test sets.
The second S2ST dataset is from VoxPopuli~\cite{wang-etal-2021-voxpopuli}, which contains speech from European parliament plenary sessions and the oral interpretation (dubbed as ``\textit{S2ST-real}''). We use the same training data in~\cite{lee2021textless} and apply the speech normalizer from the 1-hr setup on the target speech. 

The wav2vec 2.0 speech encoder and unit mBART are pre-trained on \textbf{unlabeled speech data}. For VoxPopuli~\cite{wang-etal-2021-voxpopuli}, we remove utterances from the year 2012 and before to avoid overlap with the Europarl-ST~\cite{iranzo2020europarl} dev and test data. For data augmentation, we use all the \textbf{ASR data} and the text transcriptions of the source speech in the S2T datasets to train the ASR models, and all the \textbf{parallel text data} to train the MT models. 

\vspace{-2mm}
\subsection{Model setup}
\label{sec:model_setup}
We use the multilingual HuBERT (mHuBERT) model, k-means model and unit-based HiFi-GAN vocoder from~\cite{lee2021textless}\cref{fn:textless}, to encode target speech into a vocabulary of 1000 units.
The mHuBERT and the k-means models are learned from the combination of En, Es and French unlabeled speech data from VoxPopuli~\cite{wang-etal-2021-voxpopuli}, while we use them to encode En and Es target speech only.

We train the Conformer wav2vec 2.0 speech encoder with the \textsc{LARGE} configuration~\cite{baevski2020wav2vec} using Libri-light~\cite{kahn2020libri} for En and VoxPopuli~\cite{wang-etal-2021-voxpopuli} for Es, respectively, for 200k updates with a batch size of 19.4-hr for Es and 14.7-hr for En.
We train the unit mBART with the \textsc{LARGE} configuration~\cite{vaswani2017attention} using the combination of all En and Es unlabeled speech for 500k updates with $\lambda=10$ and $p=0.3$, and we do not use the sentence permutation noise.
During finetuning, we tune the hyper-parameters including learning rate ([5e-5,~1e-4]), dropout ([0.1,~0.3]), label smoothing ([0,~0.3]) and also encoder specific ones namely mask channel length ([10,~32,~64]), mask probability ([0.1,~0.5]), channel mask probability ([0.1,~0.5]), layer drop ([0,~0.3]) and number of updates to freeze the wav2vec 2.0 encoder ([0,~5k]) on the dev sets.

\vspace{-2mm}
\subsection {Baselines}
We build two cascaded baselines, ASR+MT+TTS and S2T+TTS\cref{fn:tts}, and two supervised S2UT baselines:
\begin{enumerate*}[leftmargin=*]
    \item \textbf{ASR}: The En ASR model is finetuned with CTC from the Conformer wav2vec 2.0 model. We apply the same training for Es but find that a supervised ASR model with the \texttt{s2t\_transformer\_l} architecture in \textsc{fairseq} is better.
    \item \textbf{MT}: As the ASR models are trained with normalized text (e.g.~lowercase, digits in spoken form, etc.), we apply text normalization on both source and target texts as well to train Es-En and En-Es MT models.
    We use the \texttt{transformer\_wmt\_en\_de\_big} architecture in \textsc{fairseq}. The MT models are also used in data augmentation.
    \item \textbf{S2T}: The S2T model consists of the pre-trained Conformer wav2vec 2.0 encoder and a randomly initialized text decoder with 6 Transformer layers, 8 attention heads, 256 embedding size and 2048 FFN embedding size, and is trained on the S2T datasets without multitask learning. 
    \item \textbf{Supervised S2UT}: We follow the same model configuration in~\cite{lee2021direct} to train Transformer-based S2UT models and explore both without and with multitask learning. For the latter we include two auxiliary tasks that use character sequences from source and target text transcripts as targets.
\end{enumerate*}

\begin{table*}[t!]
  \caption{Dev / test BLEU on all the datasets included in the ``\textit{S2ST-syn}'' data. All S2UT systems are decoded with beam size 10. MOS is reported with 95\% confidence interval. (w2v2-L: wav2vec 2.0 \textsc{large})}
  \label{tab:main_results}
  \centering
\resizebox{\linewidth}{!}{
  \begin{tabular}{c|lcc|c|ccc|c}
    \hline
& & \multicolumn{3}{c|}{En-Es} & \multicolumn{4}{c}{Es-En} \\
& & \multicolumn{2}{c|}{BLEU} & MOS & \multicolumn{3}{c|}{BLEU} & MOS \\
ID &  & Europarl-ST & MuST-C & combined & CoVoST-2 & Europarl-ST & mTEDx & combined  \\
\hline
\multicolumn{8}{l}{\textbf {Cascaded systems:}} \\
1 & S2T (w2v2-L)+TTS & 33.0 / 32.6 & 30.3 / 30.1 & 3.80 $\pm$ 0.12 & 25.9 / 28.4 & 26.9 / 23.6 & 25.3 / 21.5 & 3.53 $\pm$ 0.14 \\
2 & ASR+MT+TTS & 28.9 / 28.8  & 36.4 / 34.2 & - & 37.3 / 33.8 & 33.3 / 29.1 & 29.3 / 32.4 & - \\
\hline 
\multicolumn{8}{l}{\textbf {S2UT systems without pre-training:}} \\
3 & S2UT (w/o multitask)~\cite{lee2021direct} & 23.8 / 24.0 &  25.0 / 23.3 & - & 0.0 / 0.0 & 0.0 / 0.0 & 0.1 / 0.0 & - \\
4 & S2UT (w/ multitasks)~\cite{lee2021direct} & 25.5 / 25.8 & 26.3 / 24.3 & 3.97 $\pm$ 0.09 &  20.6 / 22.7 & 20.4 / 18.0 & 20.2 / 16.9 & 3.26 $\pm$ 0.09   \\
\hline 
\multicolumn{8}{l}{\textbf {S2UT systems with model pre-training:}} \\
5 & w2v2-L    & 30.8 / 31.0 & 31.1 / 30.3 & 3.35 $\pm$ 0.15 & 24.4 / 27.0 & 24.2 / 21.5 & 24.3 / 21.0  & 3.15 $\pm$ 0.14 \\
6 & w2v2-L + mBART (LNA-E) & 30.1 / 30.4 & 31.0 / 28.2  & - &  24.4 / 27.1 & 24.0 / 21.4 & 23.6 / 21.1 & - \\
7 & w2v2-L + mBART (LNA-D) & \textbf{32.2} / \textbf{32.5} & \textbf{32.6} / \textbf{30.8} & 4.06 $\pm$ 0.10 & \textbf{27.3} / \textbf{30.2} & \textbf{29.0} / \textbf{26.4} & \textbf{29.6} / \textbf{25.2} & 2.81 $\pm$ 0.16 \\
8 & w2v2-L + mBART (LNA-E,D) & 30.6 / 31.0 & 31.3 / 29.3 & - & 26.8 / 29.6 & 27.6 / 25.2 & 24.7 / 22.3  & - \\
9 & w2v2-L + mBART (full) & 31.4 / 30.8 & 31.2 / 30.5 & - & 27.3 / 30.1 & 27.0 / 24.4 & 26.6 / 24.2 &  - \\
\hline
\multicolumn{8}{l}{\textbf {S2UT systems with model pre-training and data augmentation:}} \\
10 & w2v2-L + mBART (LNA-D) & \textbf{32.9} / \textbf{32.7} & \textbf{33.5} / \textbf{32.1} & - & \textbf{30.7} / \textbf{33.5} & \textbf{31.4} / \textbf{28.6} & \textbf{33.4} / \textbf{29.1}  & - \\
\hline
11 & Synthetic target & 89.4 / - & 86.3 / - & - & 81.3 / - & 85.8 / - & 88.4 / - & - \\
  \end{tabular}
}
\vspace{-0.3cm}
\end{table*}

\begin{table}[th!]
  \caption{Dev and test BLEU on Europarl-ST from models trained with the ``\textit{S2ST-real}'' data. }
  \label{tab:real_data}
  \centering
  \resizebox{\linewidth}{!}{
  \begin{tabular}{lcccc}
    \hline
& \multicolumn{2}{c}{En-Es} & \multicolumn{2}{c}{Es-En} \\
 & dev & test & dev & test  \\
\hline
 w/ multitask~\cite{lee2021textless} & - & 21.8 & - & 18.8 \\
 w2v2-L + mBART (LNA-D)  & 25.7  & 26.0  & 25.7  & 23.8 \\
\end{tabular}}
\end{table}

\begin{table}[th!]
  \caption{Dev and test BLEU from models trained with 10-hr, 50-hr, and 100-hr subsets of ``\textit{S2ST-syn}''. Results are averaged from the multiple datasets of each language direction.}
  \label{tab:low_res}
  \centering
\resizebox{\linewidth}{!}{
  \begin{tabular}{l|c|ccccc}
    \hline
& \multirow{2}{*}{hours} & \multicolumn{2}{c}{En-Es} & \multicolumn{2}{c}{Es-En} \\
 & & dev & test & dev & test \\
\hline
w/ multitask~\cite{lee2021direct} & 10 & 0.5 & 0.5 & 0.8 & 0.7 \\
w2v2-L + mBART (LNA-D) & 10 & 0.3 &  0.3 & 1.2 & 1.1 \\
\hline
w/ multitask~\cite{lee2021direct} & 30 & 7.6 & 7.9 & 8.2 & 7.4 \\
w2v2-L + mBART (LNA-D) & 30 & 10.5 &  11.3 & 18.9 & 17.7 \\
\hline
w/ multitask~\cite{lee2021direct} & 50 & 9.7 & 10.1 & 11.9 & 11.1\\
w2v2-L + mBART (LNA-D) & 50 & 19.3 &  18.9 & 22.4 & 21.4 \\
\hline
w/ multitask~\cite{lee2021direct} & 100 & 12.1 & 13.1 & 15.6 & 14.6 \\
w2v2-L + mBART (LNA-D) & 100 & 26.8 & 26.1  & 27.0 & 26.2 \\
\end{tabular}}
\vspace{-0.5cm}
\end{table}

\vspace{-2mm}
\subsection{Evaluation}
To evaluate the translation quality, we use open-sourced ASR models\footnote{En: \url{https://huggingface.co/facebook/wav2vec2-large-960h-lv60-self}, Es: \url{https://huggingface.co/jonatasgrosman/wav2vec2-large-xlsr-53-spanish}} to transcribe the audios and compute BLEU scores using \textsc{SacreBLEU}~\cite{post2018call}.
The reference text is normalized to lowercase, punctuation is removed, 
digits are converted to spoken forms, and all words in parentheses like ``(Applause)'' or ``(Music)'' are removed. We do not consider samples with empty translation after text normalization.
To evaluate the naturalness of the speech output, we collect mean opinion scores (MOS) on a scale of 1 (the worst) to 5 (the best) from human listening tests on a set of 200 utterances randomly sampled across all the test sets for each system, and each sample is rated by 7 raters. 

\subsection {Results}
\subsubsection{S2ST with model pre-training}
Table~\ref{tab:main_results} shows results from models trained with ``\textit{S2ST-syn}'' data.
We also provide BLEU from the synthetic targets (11) to demonstrate the impact of ASR errors.
First, we see that without multitask, the supervised Es-En S2UT model (3) cannot converge properly with the combined 196-hr training set, while multitask learning helps model training (4).

Next, we see that with a pre-trained wav2vec 2.0 encoder and a randomly initialized decoder, we can achieve an average of 5.6 BLEU gain on En-Es test sets and 4.0 BLEU for Es-En (4 vs.~5).
As we incorporate a unit mBART decoder, we find that \textbf{LNA-D} is the most effective finetuning strategy, yielding an average of 6.6 BLEU gain on En-Es test sets and 8.1 BLEU on Es-En compared to multitask learning (4 vs.~7).
Our best En-Es S2ST model performs on par with the S2T+TTS baseline, and the Es-En S2ST model outperforms the cascaded system by 2.8 BLEU (1 vs.~7). Though the S2T system can be further improved with text-based pre-training, this is beyond our scope.

Further incorporating weakly supervised training data from ASR speech can bring +0.7 BLEU on En-Es and +3.1 BLEU on Es-En (7 vs.~10). We can also compare this system with the three-stage cascaded system, as they incorporate information from the same amount of supervised ASR and MT data. We see that the En-Es direct system can outperform ASR+MT+TTS, while there is a 1.4 BLEU gap for Es-En systems (2 vs.~10).

For MOS, Table~\ref{tab:main_results} shows that the En-Es S2UT system produces more natural speech than Es TTS does (1 vs.~7), while the quality of the Es-En S2UT output is much worse. Note that the naturalness of the output speech is mainly controlled by the unit vocoder, and we use the models from~\cite{lee2021textless} without finetuning.

One advantage of self-supervised pre-training is that it only uses speech data and can work for unwritten languages.
We finetune on ``\textit{S2ST-real}'' data and compares with~\cite{lee2021textless} that incorporates an auto-encoding auxiliary task to predict the discrete units extracted from the source speech. We use dev and test sets from Europarl-ST, since it's in the same domain as the ``\textit{S2ST-real}'' data.
Table~\ref{tab:real_data} shows a 4.2 and 5 BLEU gain for En-Es and Es-En with \textbf{LNA-D}.

Finally, we study the effect of pre-training, with a decreasing amount of parallel data for finetuning by randomly sampling a subset from the ``\textit{S2ST-syn}'' training data. Table~\ref{tab:low_res} compares \textbf{LNA-D} finetuning with supervised S2UT models trained with multitask learning. When training on less than 100 hours of data, we reduce the supervised S2UT model to 8 encoder layers and 4 decoder attention heads.
We see an average 3.4-13.0 BLEU gain from pre-training and \textbf{LNA-D} finetuning for both language directions when training with more than 30 hours of data.
Pre-training and finetuning with 50-hr data can already outperform supervised systems trained with 100-hr data. However, when the amount of parallel data is decreased to 10 hrs, both multitask learning and pre-training cannot work.

\vspace{-2mm}
\subsubsection{Model variations}
\label{sec:model_variation}
\vspace{-2mm}
First, we examine Conformer vs.~Transformer wav2vec 2.0. We perform finetuning with the pre-trained encoder and a randomly initialized unit decoder using Es-En ``\textit{S2ST-syn}'' data and see that Conformer wav2vec 2.0-\textsc{LARGE} model gives an average 4.6 BLEU gain compared with Transformer \textsc{LARGE} model.

Next, we study how $\lambda$ and $p$ affect unit mBART by training the model for 300k updates and finetuning on a unit-to-unit translation task, where both source and target speech are converted to \textit{reduced} discrete unit sequences. From Table~\ref{tab:mbart_ablation}, we do not see large difference except when $p=0.3$ and $\lambda=5$.

\begin{table}[th] 
\vspace{-3mm}
  \caption{Average BLEU on ``\textit{S2ST-syn}'' En-Es dev sets with respect to hyper-parameters used in unit mBART training.} 
  \label{tab:mbart_ablation}
  \centering
  \begin{tabular}{cc|ccc}
    \hline
    & & \multicolumn{3}{c}{$\lambda$} \\
    \multicolumn{2}{c|}{En-Es dev BLEU} & 5 & 10 & 15 \\
    \hline
    \multirow{3}{*}{$p$} & 0.3 & 22.7 & 23.6 & 23.2 \\
    & 0.5 & 23.2 & 23.1 & 23.5 \\
    & 0.7 & 23.4 & 23.5 & 23.3 \\
  \end{tabular}
\vspace{-5mm}
\end{table}

\vspace{-2mm}
\section{Conclusions}
In this work, we study self-supervised pre-training and data augmentation for direct S2ST models. We take advantage of an S2UT framework that encodes target speech into discrete representations, apply wav2vec 2.0 speech encoder and unit mBART decoder pre-training and perform partial finetuning.
Experiments under various setups including synthetic and real target speech and low-resource all verify the effectiveness of the approach.
We also show that applying MT to create weakly supervised data from speech in the source language can be further combined with pre-training to improve model performance.  


\section{Acknowledgements}
We would like to thank Justine Kao and Brian Bui for the help on MOS evaluation.

\bibliographystyle{IEEEtran}
\bibliography{paper}

\begin{thebibliography}{10}
\providecommand{\url}[1]{#1}
\csname url@samestyle\endcsname
\providecommand{\newblock}{\relax}
\providecommand{\bibinfo}[2]{#2}
\providecommand{\BIBentrySTDinterwordspacing}{\spaceskip=0pt\relax}
\providecommand{\BIBentryALTinterwordstretchfactor}{4}
\providecommand{\BIBentryALTinterwordspacing}{\spaceskip=\fontdimen2\font plus
\BIBentryALTinterwordstretchfactor\fontdimen3\font minus
  \fontdimen4\font\relax}
\providecommand{\BIBforeignlanguage}[2]{{%
\expandafter\ifx\csname l@#1\endcsname\relax
\typeout{** WARNING: IEEEtran.bst: No hyphenation pattern has been}%
\typeout{** loaded for the language `#1'. Using the pattern for}%
\typeout{** the default language instead.}%
\else
\language=\csname l@#1\endcsname
\fi
#2}}
\providecommand{\BIBdecl}{\relax}
\BIBdecl

\bibitem{jia2019direct}
Y.~Jia, R.~J. Weiss \emph{et~al.}, ``Direct speech-to-speech translation with a
  sequence-to-sequence model,'' \emph{Proc. Interspeech}, 2019.

\bibitem{tjandra2019speech}
A.~Tjandra, S.~Sakti \emph{et~al.}, ``Speech-to-speech translation between
  untranscribed unknown languages,'' in \emph{ASRU}, 2019.

\bibitem{zhang2020uwspeech}
C.~Zhang, X.~Tan \emph{et~al.}, ``{UWS}peech: Speech to speech translation for
  unwritten languages,'' \emph{arXiv:2006.07926}, 2020.

\bibitem{lee2021direct}
A.~Lee, P.-J. Chen \emph{et~al.}, ``Direct speech-to-speech translation with
  discrete units,'' \emph{arXiv:2107.05604}, 2021.

\bibitem{jia2021translatotron}
Y.~Jia, M.~T. Ramanovich \emph{et~al.}, ``Translatotron 2: Robust direct
  speech-to-speech translation,'' \emph{arXiv:2107.08661}, 2021.

\bibitem{lee2021textless}
A.~Lee, H.~Gong \emph{et~al.}, ``Textless speech-to-speech translation on real
  data,'' \emph{arXiv:2112.08352}, 2021.

\bibitem{lavie1997janus}
A.~Lavie, A.~Waibel \emph{et~al.}, ``{JANUS-III}: Speech-to-speech translation
  in multiple languages,'' in \emph{ICASSP}, 1997.

\bibitem{nakamura2006atr}
S.~Nakamura, K.~Markov \emph{et~al.}, ``The {ATR} multilingual speech-to-speech
  translation system,'' \emph{IEEE Transactions on Audio, Speech, and Language
  Processing}, vol.~14, no.~2, pp. 365--376, 2006.

\bibitem{lakhotia2021generative}
K.~Lakhotia, E.~Kharitonov \emph{et~al.}, ``Generative spoken language modeling
  from raw audio,'' \emph{arXiv:2102.01192}, 2021.

\bibitem{polyak2021speech}
A.~Polyak, Y.~Adi \emph{et~al.}, ``Speech resynthesis from discrete
  disentangled self-supervised representations,'' \emph{arXiv:2104.00355},
  2021.

\bibitem{berard2018end}
A.~B{\'e}rard, L.~Besacier \emph{et~al.}, ``End-to-end automatic speech
  translation of audiobooks,'' in \emph{ICASSP}, 2018.

\bibitem{bahar2019comparative}
P.~Bahar, T.~Bieschke \emph{et~al.}, ``A comparative study on end-to-end speech
  to text translation,'' in \emph{ASRU}, 2019.

\bibitem{stoian2020analyzing}
M.~C. Stoian, S.~Bansal \emph{et~al.}, ``Analyzing asr pretraining for
  low-resource speech-to-text translation,'' in \emph{ICASSP}, 2020.

\bibitem{li2021multilingual}
X.~Li, C.~Wang \emph{et~al.}, ``Multilingual speech translation from efficient
  finetuning of pretrained models,'' in \emph{ACL}, 2021.

\bibitem{weiss2017sequence}
R.~J. Weiss, J.~Chorowski \emph{et~al.}, ``Sequence-to-sequence models can
  directly translate foreign speech,'' \emph{Proc. Interspeech}, 2017.

\bibitem{jia2019leveraging}
Y.~Jia, M.~Johnson \emph{et~al.}, ``Leveraging weakly supervised data to
  improve end-to-end speech-to-text translation,'' in \emph{ICASSP}, 2019.

\bibitem{pino2020self}
J.~Pino, Q.~Xu \emph{et~al.}, ``Self-training for end-to-end speech
  translation,'' \emph{Proc. Interspeech}, 2020.

\bibitem{wang2021large}
C.~Wang, A.~Wu \emph{et~al.}, ``Large-scale self-and semi-supervised learning
  for speech translation,'' \emph{arXiv:2104.06678}, 2021.

\bibitem{kano2021transformer}
T.~Kano, S.~Sakti \emph{et~al.}, ``Transformer-based direct speech-to-speech
  translation with transcoder,'' in \emph{SLT}, 2021.

\bibitem{baevski2020wav2vec}
A.~Baevski, Y.~Zhou \emph{et~al.}, ``wav2vec 2.0: A framework for
  self-supervised learning of speech representations,'' \emph{Neurips},
  vol.~33, pp. 12\,449--12\,460, 2020.

\bibitem{liu2020multilingual}
Y.~Liu, J.~Gu \emph{et~al.}, ``Multilingual denoising pre-training for neural
  machine translation,'' \emph{Transactions of the Association for
  Computational Linguistics}, vol.~8, pp. 726--742, 2020.

\bibitem{yang2021superb}
S.-w. Yang, P.-H. Chi \emph{et~al.}, ``{SUPERB}: Speech processing universal
  performance benchmark,'' \emph{arXiv:2105.01051}, 2021.

\bibitem{hsu2021hubert}
W.-N. Hsu, B.~Bolte \emph{et~al.}, ``{H}u{BERT}: Self-supervised speech
  representation learning by masked prediction of hidden units,''
  \emph{arXiv:2106.07447}, 2021.

\bibitem{chen2021wavlm}
S.~Chen, C.~Wang \emph{et~al.}, ``Wav{LM}: Large-scale self-supervised
  pre-training for full stack speech processing,'' \emph{arXiv:2110.13900},
  2021.

\bibitem{ao2021speecht5}
J.~Ao, R.~Wang \emph{et~al.}, ``Speech{T}5: Unified-modal encoder-decoder
  pre-training for spoken language processing,'' \emph{arXiv:2110.07205}, 2021.

\bibitem{kharitonov2021text}
E.~Kharitonov, A.~Lee \emph{et~al.}, ``Text-free prosody-aware generative
  spoken language modeling,'' \emph{arXiv:2109.03264}, 2021.

\bibitem{kreuk2021textless}
F.~Kreuk, A.~Polyak \emph{et~al.}, ``Textless speech emotion conversion using
  decomposed and discrete representations,'' \emph{arXiv:2111.07402}, 2021.

\bibitem{tjandra2018machine}
A.~Tjandra, S.~Sakti \emph{et~al.}, ``Machine speech chain with one-shot
  speaker adaptation,'' \emph{Proc. Interspeech}, 2018.

\bibitem{kahn2020self}
J.~Kahn, A.~Lee \emph{et~al.}, ``Self-training for end-to-end speech
  recognition,'' in \emph{ICASSP}, 2020.

\bibitem{hayashi2018back}
T.~Hayashi, S.~Watanabe \emph{et~al.}, ``Back-translation-style data
  augmentation for end-to-end asr,'' in \emph{SLT}, 2018.

\bibitem{vaswani2017attention}
A.~Vaswani, N.~Shazeer \emph{et~al.}, ``Attention is all you need,'' in
  \emph{Neurips}, 2017, pp. 5998--6008.

\bibitem{gulati2020conformer}
A.~Gulati, J.~Qin \emph{et~al.}, ``Conformer: Convolution-augmented
  {T}ransformer for speech recognition,'' \emph{Proc. Interspeech}, 2020.

\bibitem{ott2019fairseq}
M.~Ott, S.~Edunov \emph{et~al.}, ``fairseq: A fast, extensible toolkit for
  sequence modeling,'' in \emph{Proc. NAACL-HLT 2019: Demonstrations}, 2019.

\bibitem{wang2020fairseqs2t}
C.~Wang, Y.~Tang \emph{et~al.}, ``fairseq {S2T}: Fast speech-to-text modeling
  with fairseq,'' in \emph{AACL: System Demonstrations}, 2020.

\bibitem{wang2020covost}
C.~Wang, A.~Wu \emph{et~al.}, ``Covost 2 and massively multilingual
  speech-to-text translation,'' \emph{arXiv:2007.10310}, 2020.

\bibitem{iranzo2020europarl}
J.~Iranzo-S{\'a}nchez, J.~A. Silvestre-Cerd{\`a} \emph{et~al.}, ``Europarl-st:
  A multilingual corpus for speech translation of parliamentary debates,'' in
  \emph{ICASSP}, 2020.

\bibitem{salesky2021mtedx}
E.~Salesky, M.~Wiesner \emph{et~al.}, ``Multilingual tedx corpus for speech
  recognition and translation,'' in \emph{Proc. of Interspeech}, 2021.

\bibitem{cattoni2021must}
R.~Cattoni, M.~A. Di~Gangi \emph{et~al.}, ``Must-c: A multilingual corpus for
  end-to-end speech translation,'' \emph{Computer Speech \& Language}, vol.~66,
  p. 101155, 2021.

\bibitem{wang2021voxpopuli}
C.~Wang, M.~Riviere \emph{et~al.}, ``Voxpopuli: A large-scale multilingual
  speech corpus for representation learning, semi-supervised learning and
  interpretation,'' \emph{arXiv:2101.00390}, 2021.

\bibitem{pratap2020mls}
V.~Pratap, Q.~Xu \emph{et~al.}, ``{MLS}: A large-scale multilingual dataset for
  speech research,'' \emph{arXiv:2012.03411}, 2020.

\bibitem{ardila2019common}
R.~Ardila, M.~Branson \emph{et~al.}, ``Common voice: A massively-multilingual
  speech corpus,'' \emph{arXiv:1912.06670}, 2019.

\bibitem{panayotov2015librispeech}
V.~Panayotov, G.~Chen \emph{et~al.}, ``Librispeech: an asr corpus based on
  public domain audio books,'' in \emph{ICASSP}, 2015.

\bibitem{hernandez2018ted}
F.~Hernandez, V.~Nguyen \emph{et~al.}, ``{TED-LIUM} 3: twice as much data and
  corpus repartition for experiments on speaker adaptation,'' in
  \emph{International conference on speech and computer}.\hskip 1em plus 0.5em
  minus 0.4em\relax Springer, 2018, pp. 198--208.

\bibitem{kahn2020libri}
J.~Kahn, M.~Riviere \emph{et~al.}, ``Libri-light: A benchmark for {ASR} with
  limited or no supervision,'' in \emph{ICASSP}, 2020.

\bibitem{lison-etal-2018-opensubtitles2018}
P.~Lison, J.~Tiedemann \emph{et~al.}, ``{O}pen{S}ubtitles2018: Statistical
  rescoring of sentence alignments in large, noisy parallel corpora,'' in
  \emph{{LREC}}, 2018.

\bibitem{ziemski2016united}
M.~Ziemski, M.~Junczys-Dowmunt \emph{et~al.}, ``The united nations parallel
  corpus v1. 0,'' in \emph{LREC}, 2016.

\bibitem{skadins-etal-2014-billions}
R.~Skadi{\c{n}}{\v{s}}, J.~Tiedemann \emph{et~al.}, ``Billions of parallel
  words for free: Building and using the {EU} bookshop corpus,'' in
  \emph{LREC}, 2014.

\bibitem{koehn2005europarl}
P.~Koehn, ``Europarl: A parallel corpus for statistical machine translation,''
  in \emph{Proc. machine translation summit x: papers}, 2005.

\bibitem{wolk2014building}
K.~Wo{\l}k and K.~Marasek, ``Building subject-aligned comparable corpora and
  mining it for truly parallel sentence pairs,'' \emph{Procedia Technology},
  vol.~18, pp. 126--132, 2014.

\bibitem{reimers-2020-multilingual-sentence-bert}
N.~Reimers and I.~Gurevych, ``Making monolingual sentence embeddings
  multilingual using knowledge distillation,'' in \emph{EMNLP}, 2020.

\bibitem{likhomanenko2020rethinking}
T.~Likhomanenko, Q.~Xu \emph{et~al.}, ``Rethinking evaluation in asr: Are our
  models robust enough?'' \emph{arXiv:2010.11745}, 2020.

\bibitem{chan2021speechstew}
W.~Chan, D.~Park \emph{et~al.}, ``Speechstew: Simply mix all available speech
  recognition data to train one large neural network,''
  \emph{arXiv:2104.02133}, 2021.

\bibitem{wang-etal-2021-voxpopuli}
C.~Wang, M.~Riviere \emph{et~al.}, ``{V}ox{P}opuli: A large-scale multilingual
  speech corpus for representation learning, semi-supervised learning and
  interpretation,'' in \emph{ACL}, 2021.

\bibitem{post2018call}
M.~Post, ``A call for clarity in reporting {BLEU} scores,'' in \emph{Proc. the
  Third Conference on Machine Translation: Research Papers}, 2018, pp.
  186--191.

\end{thebibliography}

\appendix
\section{Model Training Hyper-parameters}
\subsection{Wav2vec 2.0}

To train the wav2vec 2.0 models, we use the \texttt{wav2vec2\_conformer\_large\_librivox} configuration defined in \textsc{fairseq}. The model contains 24 Conformer blocks with model dimension 1024, inner dimension 4096, 16 attention heads and a total of 620.5M parameters. We use a total batch size of 14.7-hr for En and 19.4-hr for Es, and dropout 0.1. We use Adam with $\beta_{1}=0.9, \beta_{2}=0.98, \epsilon=10^{-6}$, weight decay of 0.1 and learning rate 0.005, and apply polynomial decay learning rate schedule with 32k warmup steps. We started the training with fp16 initially but switched to fp32 when we encountered NaNs in the back propagation. Rest of the hyper-parameters are listed in Table~\ref{tab:wav2vec2_hp}.

\begin{table}[th]
\caption{\label{tab:wav2vec2_hp} Hyper-parameters used in wav2vec 2.0 training}
  \centering
    \begin{tabular}{l|c|c}
    \hline
      hyper-parameter & En & Es \\
    \hline
      max tokens per GPU & 550,000 & 700,000 \\
      \#GPUs & 256 & 400 \\
      max positions & 5000 & 5000 \\
      update freq & 6 & 4 \\
      max updates & 220,000 & 188,000 \\
    \hline
    \end{tabular}
\end{table}

\subsection{Unit mBART}
We use the \texttt{mbart\_large} architecture defined in \textsc{fairseq} to train the unit mBART.
The model contains 12 Transformer layers in both encoder and decoder, embedding size 1024, feed-forward network (FFN) dimension 4096, 16 attention heads and a total of 353M parameters.
Different from original text mBART, we do not apply sentence permutation noise in the denoising task, and we only apply masking noise.
We use Adam with $\beta_{1}=0.9, \beta_{2}=0.999, \epsilon=10^{-6}$ and learning rate 0.0003, and apply polynomial decay learning rate schedule with 10000 warmup steps.
The model is trained with dropout 0.1, mask probability 0.3, $\lambda=10$ for 500k steps. We use 1024 max tokens with 64 GPUs and update frequency 9.

\subsection{Supervised S2UT baselines}
To train the supervised S2UT models with multitasks, we use the \texttt{s2ut\_transformer\_fisher} configuration defined in \textsc{fairseq}. We use Adam with $\beta_{1}=0.9, \beta_{2}=0.98, \epsilon=10^{-8}$, weight decay of 0.1 and learning rate 0.0005, and apply inverse square root learning rate schedule with 10k warmup steps. For En-Es, we use a max tokens value of 3000, dropout value of 0.1 and 16 GPUs. For Es-En, we use 4500 max tokens value, dropout 0.3 and 8 GPUs.

For auxiliary tasks, we have a Transformer decoder on the sixth layer of the encoder for source character prediction and a Transformer decoder on the eighth layer of the encoder for target character prediction. Each of the multitasks has a loss weight of 8. The decoders have 2 Transformer layers with 256 embedding size and FFN dimension of 2048.

\subsection{Finetuning}
\label{sec:finetune_hp}
We use a dropout value of 0.1 and label smoothing value of 0.2 for both language directions. We use encoder layerdrop value of 0.1 for En-Es and 0.2 for Es-En. For the decoder, we use the same embedding parameters for the input and output. We use Adam with $\beta_{1}=0.9, \beta_{2}=0.98, \epsilon=10^{-6}$ and apply inverse square root learning rate schedule with 10k warmup steps. We finetune the models for 15k updates with ``\textit{S2ST-syn}'' data and 25k updates when incorporating the weakly supervised data.
We use max tokens of 3500 for En-Es and 4000 for Es-En, set the update frequency to 15 and train on 16 GPUs. The learning rate and the number of updates we freeze the wav2vec 2.0 encoder in the beginning for each of the setups is listed in the Table~\ref{tab:wav2vec2_mbart_hp}.

\begin{table*}[th!]
\caption{\label{tab:wav2vec2_mbart_hp} Hyper-parameters used in wav2vec 2.0 and mBART finetuning }
  \centering
  \begin{tabular}{lc|cc|cc}
    \hline
& \#Parameters & \multicolumn{2}{c}{En-Es} & \multicolumn{2}{c}{Es-En} \\
& Finetuned & learning &  encoder & learning &   encoder \\
& (Millions) & rate &  freeze steps & rate & freeze steps \\
\hline
w2v2-L & 628.9 & 0.0005  & 5000 & 0.0008  & 0\\
w2v2-L + mBART (LNA-E) & 335.1 & 0.0005  & 0 & 0.0008 &  0\\
w2v2-L + mBART (LNA-D) & 725.7 & 0.0005  & 5000 & 0.001 &  0\\
w2v2-L + mBART (LNA-E,D) & 233.3 & 0.0005  & 5000 & 0.0008 &   0\\
w2v2-L + mBART (full) & 827.4 & 0.0005  & 0 & 0.001 &  0 \\
\hline
\multicolumn{5}{l}{\textbf {Data augmentation:}} \\
w2v2-L + mBART (LNA-D) & 725.7 & 0.0005 &  5000 & 0.0005 &  5000\\
\end{tabular}
\end{table*}

\subsection{Low-resource setup}
Table~\ref{tab:lowres_dur} lists the statistics of the training data used in the low-resource setup.
We use the same set of hyper-parameters for the wav2vec 2.0 and mBART finetuning on the low-resource setup as described in Sec.~\ref{sec:finetune_hp} except for the ones listed in Table~\ref{tab:wav2vec2_mbart_LR_hp}.
Finally, Table~\ref{tab:LR results} lists the BLEU scores on each dataset under the low-resource setup, whose average values are presented in Table~\ref{tab:low_res}.

\begin{table*}[th!]
\caption{\label{tab:lowres_dur} Duration (hours) of the training data sampled from each dataset in the low-resource setup  }
  \centering
  \begin{tabular}{c|cc|ccc}
    \hline
 & \multicolumn{2}{c}{En-Es} & \multicolumn{3}{c}{Es-En} \\
total hours & Europarl-ST & MuST-C  & CoVoST-2 & Europarl-ST & mTEDx   \\
\hline
10 & 1.4 & 8.6 & 5.8 & 1.0 & 3.2 \\
30 & 4.0 & 26.0 & 17.2 & 3.0 & 9.9 \\
50 & 6.7 & 43.4 & 28.7 & 5.2 & 16.1 \\
100 & 13.5 & 86.5 & 57.4 & 10.2 & 32.3 \\
\end{tabular}
\end{table*}

\begin{table*}[th!]
\caption{\label{tab:wav2vec2_mbart_LR_hp} Hyper-parameters used in low-resource LNA-D finetuning experiments }
  \centering
  \begin{tabular}{c|cc|cc}
    \hline
 & \multicolumn{2}{c}{En-Es} & \multicolumn{2}{c}{Es-En} \\
\multirow{2}{*}{hours} & learning  & encoder & learning &  encoder \\
& rate  & freeze steps & rate & freeze steps \\
\hline
10 & 0.0001  & 5000 & 0.0008  & 0 \\
30 & 0.0001 & 5000 & 0.001  & 0 \\
50 & 0.0001  & 5000 & 0.0008  & 0 \\
100 & 0.0001  & 5000 & 0.0008  & 0\\

\end{tabular}
\end{table*}

\begin{table*}[t!]
  \caption{Dev / test BLEU on all the datasets included in the ``\textit{S2ST-syn}'' data in the low-resource setup. }
  \label{tab:LR results}
  \centering
  \begin{tabular}{c|lcc|ccc}
    \hline
& & \multicolumn{2}{c|}{En-Es BLEU} & \multicolumn{3}{c}{Es-En BLEU} \\
 & hours & Europarl-ST & MuST-C  & CoVoST-2 & Europarl-ST & mTEDx   \\
\hline
w/ multitask~\cite{lee2021direct} & 10 & 0.6 / 0.7 & 0.3 / 0.3 & 1.2 / 1.0 & 1.0 / 0.9 & 0.3 / 0.3 \\
w2v2-L + mBART (LNA-D) & 10 & 0.4 / 0.4 & 0.1 / 0.2 & 1.7 / 2.0 & 1.2 / 1.0 & 0.6 / 0.3\\
\hline 

w/ multitask~\cite{lee2021direct} & 30 & 8.1 / 7.9 & 7.0 / 7.8 & 9.2 / 9.5 & 8.0 / 6.8 & 7.4 / 6.0 \\
w2v2-L + mBART (LNA-D) & 30 & 11.0 / 11.6 & 10.6 / 10.9  & 19.9 / 22.3 & 18.1 / 15.9 & 18.6 / 15.1\\
\hline 

w/ multitask~\cite{lee2021direct} & 50 & 10.4 / 10.4 & 8.9 / 9.8 & 12.5 / 13.6 & 11.7 / 10.2 & 11.5 / 9.4 \\
w2v2-L + mBART (LNA-D) & 50 & 19.4 / 19.6 & 19.1 / 18.1 & 23.1 / 25.5 & 22.2 / 19.5 & 22 / 19.2\\
\hline 

w/ multitask~\cite{lee2021direct} & 100 & 13.3 / 13.6 & 11.0 / 12.6 & 16.6 / 18.2 & 15.1 / 13.1 & 15.1 / 12.6 \\
w2v2-L + mBART (LNA-D) & 100 &  27.1 / 26.6 & 26.4 / 25.6 & 26.8 / 29.6 & 27.7 / 24.9 & 26.6 / 24.1 \\
\hline 
  \end{tabular}
\end{table*}

\end{document}